\documentclass[conference]{IEEEtran}
\IEEEoverridecommandlockouts
\usepackage{cite}
\usepackage{amsmath,amssymb,amsfonts}
\usepackage{algorithmic}
\usepackage{multirow}
\usepackage{graphicx}
\usepackage{textcomp}
\usepackage{url}
\usepackage{xcolor}
\def\BibTeX{{\rm B\kern-.05em{\sc i\kern-.025em b}\kern-.08em
    T\kern-.1667em\lower.7ex\hbox{E}\kern-.125emX}}

\begin{document}
\title{Enhanced Smart Contract Reputability Analysis using Multimodal Data Fusion on Ethereum}

%\author{\IEEEauthorblockN{Paper ID: 1571108626}
%\IEEEauthorblockA{}
%}

%\maketitle

 \author{\IEEEauthorblockN{Cyrus Malik}
 \IEEEauthorblockA{\textit{Department of Artificial Intellgience} \\
 \textit{University of Malta}\\
 Msida, Malta \\
 cyrus.malik.20@um.edu.mt}
 \and
 \IEEEauthorblockN{Josef Bajada}
 \IEEEauthorblockA{\textit{Department of Artificial Intelligence} \\
 \textit{University of Malta}\\
 Msida, Malta \\
 josef.bajada@um.edu.mt}
 \and
 \IEEEauthorblockN{Joshua Ellul}
 \IEEEauthorblockA{\textit{Centre for DLT \&} \\
 \textit{Department of Computer Science} \\
 \textit{University of Malta}\\
 Msida, Malta \\
 joshua.ellul@um.edu.mt}
 } 
 \maketitle

\begin{abstract}
The evaluation of smart contract reputability is essential to foster trust in decentralized ecosystems. However, existing methods that rely solely on analysis of code or transactional data, offer limited insight into evolving trustworthiness. We propose a multimodal data fusion framework that integrates code feature analysis with transactional data to enhance reputability prediction. Our framework initially focuses on AI-based code analysis, utilizing GAN-augmented opcode embeddings to address class imbalance, achieving 97.67\% accuracy and a recall of 0.942 in detecting illicit contracts, surpassing traditional oversampling methods.  This forms the crux of a reputability-centric fusion strategy, where combining code and transactional data improves recall by 7.25\% over single-source models, demonstrating robust performance across validation sets. By providing a holistic view of smart contract behaviour, our approach enhances the model’s ability to assess reputability, identify fraudulent activities, and predict anomalous patterns. These capabilities contribute to more accurate reputability assessments, proactive risk mitigation, and enhanced blockchain security.
\end{abstract}


\section{Introduction}
In the rapidly evolving blockchain landscape, smart contracts have become a fundamental component, enabling automated and trustless transactions that facilitate decentralized finance, governance, and other critical applications \cite{kolvart2016smart, szabo1997formalizing}. These self-executing protocols enforce business logic transparently, eliminating the need for intermediaries. However, as the adoption of smart contracts grows, so does the importance of evaluating their reliability and reputability to mitigate risks associated with vulnerabilities, malicious exploits, erroneous code and other fraudulent activity \cite{sayeed2020smart, groce2020actual, farrugia2020detection, agarwal2021detecting, vassallo2021application}. Assessing smart contract reputability is essential to establish trust in decentralized ecosystems and prevent systemic failures.

Current reputability assessment methods predominantly rely on either code analysis or transactional data. Code analysis evaluates the contract's source code or bytecode for vulnerabilities and design flaws using tools such as SmartCheck \cite{tikhomirov2018smartcheck} and Slither \cite{groce2020actual}. Although effective in detecting inherent security issues, code analysis provides a limited view, omitting insights from real-world usage and external factors. In contrast, transactional data analysis tracks runtime behaviours such as transaction volume and user activity, providing contextual insights into contract interactions \cite{Chen2018,malik2025likely}. However, these approaches are reactive, only detecting issues after they occur, and lacks the interpretability needed to understand the underlying causes. These isolated approaches fail to capture the full spectrum of reputability, which encompasses both intrinsic code quality and dynamic behavioural patterns influenced by user interactions and external events \cite{ibrahim2021illicit, obi2023machine}.

This work addresses this gap by proposing a multimodal data fusion framework that integrates code features with dynamic transactional data to provide a comprehensive and evolving assessment of smart contract reputability. This research is guided by the following research question: Can a multimodal data fusion framework predict the evolution of smart contract reputability by integrating AI-based code analysis with transactional data?

Our framework combines opcode embeddings, augmented using Generative Adversarial Networks (GANs) to address class imbalance, with transactional data for anomaly detection. We propose the use of boosting algorithms \cite{bentejac2021comparative}, such as LightGBM, CatBoost, and XGBoost, to predict reputability based on code analysis, achieving robust performance compared to traditional oversampling methods.  Additionally, we use convolutional autoencoder-based models to fuse code and dynamic data, capturing the interplay between a contract's internal logic and its interactions \cite{agarwal2021detecting, zhao2021temporal}. This combined approach enhances anomaly detection and captures reputability shifts that single-source models often miss. The main contributions from this work are as follows:
\begin{itemize}
    \item A novel multimodal data fusion framework that integrates opcode embeddings and transactional features for comprehensive smart contract reputability analysis, incorporating anomaly detection techniques to identify reputability shifts and improve robustness.
    \item A detailed empirical evaluation demonstrating the effectiveness of boosting algorithms with GAN-augmented opcode embeddings in handling class imbalance and improving reputability classification.
    \item Publicly available datasets of labelled smart contracts and reputability scores across time, designed to support further research and address the challenge of missing reputability labels at specific timestamps.
\end{itemize}

By addressing the limitations of code analysis-only and transactional-only approaches, this research lays the groundwork for further advancements in smart contract analysis. Our framework supports the proactive identification of reputability shifts over time and anomalous behaviour, contributing to the security and reliability of blockchain ecosystems.

The remainder of this paper is organized as follows: Section II reviews related work on smart contract reputability assessment and the limitations of code analysis and transactional data approaches. Section III presents our multimodal data fusion framework, detailing the preprocessing steps, boosting algorithms, and fusion architecture. Section IV describes the experimental setup, datasets, and evaluation metrics, along with key performance comparisons. Section V discusses the limitations of our approach and future research directions. Finally, Section VI concludes the paper by summarizing our findings and emphasising the implications for blockchain security and trustworthiness.

\section{Background}

\subsection{Smart Contract Reputability}
Smart contracts are agreements embedded with code that guarantee the execution of predefined conditions, facilitating automated transactions without intermediaries and supporting trustless interactions within decentralized ecosystems \cite{szabo1997formalizing}. However, the trust in smart contracts extends beyond their self-enforcing nature—it is also determined by their reputability, which refers to the degree to which a smart contract can be relied upon to perform its intended functions securely and consistently without vulnerabilities or malicious intent. Reputability in the context of smart contracts can be assessed through multiple dimensions, including the correctness of the contract's code, its historical performance, and its adherence to best practices in coding and security standards \cite{looram2024reputation}.

In Decentralized Finance (DeFi), where smart contracts underpin critical applications such as lending, borrowing, and decentralized exchanges, reputability plays a pivotal role. While reputability is shaped by factors such as code correctness, historical performance, and security practices, events such as hacks or exploits can severely impact it. For instance, the 2016 DAO hack exploited a vulnerability in the contract’s recursive call function, resulting in the loss of approximately \$50 million worth of Ether and undermining trust in decentralized governance systems \cite{mehar2019understanding}. Despite the immutability of blockchain, which ensures that deployed contracts cannot be altered, this same property can exacerbate security risks, as any flaws in the contract become permanent.

More recent incidents highlight that even sophisticated smart contracts are not immune to devastating exploits or vulnerabilities, which can severely damage their reputability. For example, in 2022, the Nomad cross-chain interoperability protocol suffered an exploit that resulted in a \$190 million loss \cite{nomad2022exploit}. The attack stemmed from a faulty contract upgrade that inadvertently enabled unauthorized withdrawals, allowing malicious actors to systematically drain funds \cite{gucciardi2023trustless}. This high-profile failure, alongside other major DeFi breaches, underscores the critical need for robust reputability assessment mechanisms capable of detecting vulnerabilities early, continuously monitoring evolving risks, and preventing reputability degradation over time.

Traditional reputability assessments often focus on code analysis, using tools like Slither \cite{feist2019slither} to identify coding errors and vulnerabilities. However, these methods are limited in scope, as they fail to account for runtime behaviours and interactions within a decentralized ecosystem. For example, a contract that passes initial security audits may later exhibit anomalous behaviour due to unexpected external interactions or changing usage patterns. Conversely, contracts perceived as risky at launch may build reputability over time through consistent, secure operation. This dynamic nature of smart contract reputability necessitates continuous monitoring that combines code analysis with insights from transactional data to capture changes in behaviour and performance \cite{farrugia2020detection}.

Given the rise of DeFi exploits and their impact on user trust and market stability, it is imperative to move beyond isolated reputability assessments and adopt holistic frameworks that integrate code analysis and dynamic data sources. Such an approach is essential not only for safeguarding assets but also for fostering a resilient decentralized ecosystem where participants can engage confidently with smart contract-based applications.

\subsection{Limitations of Code analysis-based and Transactional Approaches}
Smart contract reputability assessments typically rely on code analysis or transactional data alone, yet both approaches have limitations that hinder a comprehensive evaluation of trustworthiness. Static code analysis tools, such as SmartCheck \cite{tikhomirov2018smartcheck} and Slither \cite{feist2019slither}, detect vulnerabilities in the source code pre-deployment, identifying issues like reentrancy flaws, unchecked calls, and arithmetic overflows. However, while effective at uncovering internal flaws in a contract’s logic, static analysis typically cannot capture external interactions or runtime behaviours (unless some form of runtime analysis or verification is used \cite{ellul2018runtime}), limiting its applicability in dynamic environments. Moreover, these tools often rely on predefined vulnerability patterns, making them less effective at identifying novel attack vectors \cite{qian2022smart}.

Code analysis often cannot account for runtime behaviours or unexpected interactions with external contracts. This limitation means that vulnerabilities arising from dynamic behaviours, such as gas usage anomalies or transaction sequencing issues, can often go undetected, potentially affecting the reputability of the smart contract if exploited \cite{he2020smart}. Moreover, analysis is also often constrained by predefined vulnerability patterns, making it less effective at identifying novel attack strategies. Recent studies \cite{deng2021ensemble} have shown that machine learning techniques, such as Long Short-term Memory (LSTMs) \cite{hochreiter1997long} and Extreme Gradient Boosting (XGBoost) \cite{chen2016xgboost}, outperform symbolic tools by detecting complex vulnerabilities through pattern recognition in opcode sequences. However, these methods still fall short when contextual runtime data is unavailable.

In contrast, transactional data-based approaches focus on runtime behaviours, analysing contract interactions, transaction volumes, and user activity to identify anomalies. These methods excel at detecting post-deployment issues, such as fraudulent behaviour and unusual fund movements. For example, the 2022 Nomad bridge exploit resulted in a \$190 million loss due to a faulty smart contract upgrade that enabled unauthorized withdrawals \cite{nomad2022exploit}. Transactional monitoring could have detected early signs of this exploit through abnormal transaction patterns. However, such approaches are inherently reactive, as they detect anomalies after they occur, and often lack the interpretability needed to pinpoint the root cause within the contract code \cite{farrugia2020detection, ibrahim2021illicit, obi2023machine, fan2022smart}.

Transactional anomaly detection models are also prone to false positives in dynamic ecosystems where legitimate transaction spikes occur due to upgrades or market events \cite{agarwal2021detecting}. For example, Ponzi scheme detection research \cite{Chen2018} demonstrates how combining behavioural features (e.g., transaction counts, participant ratios) with code-based features (e.g., opcode frequencies) improves detection accuracy. The study achieved a recall of 0.810 using XGBoost but still classified contracts only as either ``Ponzi" or ``non-Ponzi," without distinguishing the broader spectrum of reputability. This highlights a key limitation: classifying non-illicit contracts does not necessarily indicate their trustworthiness.

\subsection{Multimodal Learning for Anomaly Detection}
Multimodal data fusion integrates information from multiple data sources to create a more comprehensive and accurate understanding of a phenomenon \cite{lahat2015multimodal}. In the context of smart contracts, this involves combining code features, such as opcode embeddings and vulnerability metrics, with dynamic transactional data to capture both pre-deployment vulnerabilities and runtime behaviours. By leveraging the strengths of each data type, multimodal learning provides richer insights that single-source methods cannot, revealing patterns and relationships that may be obscured when modalities are analysed in isolation \cite{gaw2022multimodal, baltruvsaitis2018multimodal}.

In fraud detection and anomaly detection, multimodal fusion has shown promising results. For example, \cite{deng2021ensemble} proposed a model that fused source code, control flow graphs, and opcode features to detect vulnerabilities in smart contracts, achieving significant improvements in accuracy compared to single-modality models. Similarly, \cite{Chen2018} demonstrated the benefits of combining behavioural features, such as transaction frequencies and participant ratios, with opcode-based features to detect Ponzi schemes on the Ethereum blockchain. However, many of these studies framed the problem as a binary classification—fraudulent or non-fraudulent—without addressing the broader and more nuanced concept of reputability, which encompasses not just the absence of malicious intent but also adherence to best practices and consistent performance over time.

The need for multimodal fusion becomes especially apparent when considering the limitations of code and transactional data-based approaches. Code analysis identifies potential vulnerabilities in the contract’s internal logic but often fails to capture dynamic behaviours that manifest only during runtime interactions. Conversely, transactional anomaly detection captures real-time behaviours but may flag anomalous patterns without indicating whether they stem from code-level flaws or external factors. By integrating code and dynamic features, multimodal approaches bridge this gap, providing context for behavioural anomalies and enabling more accurate and interpretable assessments.

To enhance detection accuracy, feature-level fusion techniques can be employed, where features from different modalities are combined before being fed into the model. Attention mechanisms can further improve the fusion process by assigning different weights to features based on their relevance \cite{bedworth1999source}. This allows the model to prioritize critical features, leading to more precise predictions and better handling of complex patterns.

Despite these advancements, existing multimodal models often overlook key challenges in smart contract reputability analysis. One such challenge is class imbalance—datasets often contain more high-profile illicit contracts than reputable ones, which can skew predictive performance. While techniques such as GAN-based oversampling \cite{goodfellow2020generative} and XGBoost classifiers \cite{deng2021ensemble} have been applied to address this issue, many approaches fail to adapt to the evolving nature of smart contract behaviours. As contracts interact with other entities over time, their reputability can degrade or improve, highlighting the need for continuous monitoring and robust anomaly detection mechanisms.

Multimodal fusion not only improves robustness by capturing complementary information but also reduces false positives by providing a holistic view of contract behaviour. By combining code and transactional data, it becomes possible to differentiate between legitimate spikes in transaction volume (e.g., protocol upgrades) and anomalies indicative of malicious behaviour. This integration supports proactive reputability assessments, addressing key challenges in maintaining trust and security within decentralized ecosystems.

\section{Methodology}

\subsection{Datasets}
A key component of this study is the construction of a dataset comprising reputable and illicit smart contracts, specifically compiled to address the scarcity of comprehensive reputability-related data. The pseudonymous nature of cryptocurrencies \cite{nakamoto2009bitcoin, buterin2016ethereum} complicates the accurate identification of illicit contracts and hinders the development of robust reputability models. To overcome this challenge, we curated the following data as part of this study:
\begin{itemize}
    \item Reputable Smart Contracts: 3,200 contracts sourced from CoinGecko,\footnote{\url{https://www.coingecko.com/en}} a reputable cryptocurrency resource.
    \item Illicit Smart Contracts: 191 contracts retrieved from CryptoScamDB\footnote{\url{https://cryptoscamdb.org}} and prior studies \cite{farrugia2020detection}, comprising known fraudulent smart contracts.
\end{itemize}

For each smart contract, the following data types were collected using the Etherscan API\footnote{\url{https://etherscan.io/apis}}:

\begin{itemize}
    \item Source Code: Human-readable code that defines the contract's functionality and logic.
    \item Bytecode: Compiled, machine-readable code deployed on the Ethereum blockchain.
    \item Transactional Data: Records of transactions, including attributes such as block numbers, timestamps, gas usage, sender and receiver addresses.
\end{itemize}

The constructed dataset is available for download on HuggingFace\footnote{REDACTED for double-blind review} to support reproducibility and further research. Additionally, existing datasets such as the Slither Audited Smart Contracts dataset \cite{rossini2022slitherauditedcontracts} were reviewed but ultimately excluded due to their focus on vulnerability detection without reputability classifications.

\subsection{Preprocessing}
To prepare the dataset for analysis, several preprocessing steps were applied to both code and transactional data. First, the bytecode sequences of smart contracts were disassembled into opcodes using the pyevmasm library.\footnote{https://github.com/crytic/pyevmasm} To streamline the analysis and reduce complexity, each opcode was categorised to a high-level category, such as arithmetic operations, logical operations, and memory operations, thereby preserving semantic information while simplifying the sequence structure.

The simplified opcode sequences were then converted into dense vector representations using a learned embedding approach. Each opcode was represented by a 50-dimensional vector through an embedding layer implemented in TensorFlow. By averaging these embeddings across the sequence, a single feature vector was generated for each smart contract, capturing nuanced relationships in the opcode sequence and facilitating code analysis.

For transactional data, features such as the number of transactions, participant addresses, gas usage, and timestamps were extracted. These features were standardized to maintain consistent scales, ensuring that all variables contributed equally during model training. The standardization process transformed each feature to have a mean of 0 and a standard deviation of 1.

Class imbalance was addressed using oversampling techniques to balance the distribution of reputable and illicit contracts. We applied both Synthetic Minority Over-sampling Technique (SMOTE) and also Adaptive Synthetic Sampling (ADASYN) to generate synthetic samples by interpolating between existing data points, focusing on more challenging minority class examples. Additionally, GAN-based augmentation was employed to produce realistic synthetic data by learning complex patterns within the minority class. This approach enhanced the model’s ability to detect nuanced illicit behaviours, ensuring that the dataset remained representative and robust.

\subsection{Boosted Code Analysis}

To assess smart contract reputability using code features, gradient boosting algorithms were employed due to their effectiveness in handling high-dimensional data. The training features included 50-dimensional opcode embeddings derived from simplified opcode sequences. These embeddings transformed categorical opcode data into dense vector representations, effectively capturing nuanced structural patterns within contracts.

Three gradient boosting models—XGBoost \cite{chen2016xgboost}, LightGBM \cite{ke2017lightgbm}, and CatBoost \cite{dorogush2018catboost}—were selected for their ability to model complex relationships and manage imbalanced data. Hyperparameter tuning was performed using cross-validated grid search, with a 5-fold cross-validation to optimize parameters such as learning rate, tree depth, and the number of estimators. The search space included learning rates of {0.2, 0.1, 0.01}, tree depths ranging from 2 to 8, and estimators between 100 and 300. This tuning process ensured robust validation and minimized overfitting, with models evaluated across multiple configurations to identify the optimal settings for reputability classification. 

\subsection{Multimodal Data Fusion}
Once the code-level analysis was conducted, we transitioned to temporal analysis to examine how reputability evolves over time. Our focus then shifted to the characteristics of the transactions themselves. Consistent with the findings of \cite{agarwal2021detecting} and \cite{farrugia2020detection}, we concentrated primarily on normal and internal transactions since they were identified as particularly important for providing meaningful insights into the behaviour and reputability of the contracts.

We implemented a multimodal data fusion approach that integrates code features with transactional data. The code features include 50-dimensional opcode embeddings derived from disassembled and simplified bytecode sequences, which capture the intrinsic structure and vulnerability-related characteristics of the smart contract. The transactional data comprises 820,920 transactions from reputable smart contracts (post-outlier removal) and 6,761 transactions from illicit ones, aggregated at an hourly granularity to balance temporal detail with computational feasibility.

Hourly aggregation was chosen to capture short-term variations, such as transaction bursts, without oversmoothing trends as daily aggregation would. Minute-level granularity was deemed impractical due to the sheer volume of data and increased training complexity. This level of aggregation preserves critical patterns, such as spikes in gas usage and sudden changes in transaction counts, that may indicate anomalous behaviour.

The combined feature set—opcode embeddings and transactional data—was input into a convolutional autoencoder (CAE) for anomaly detection. Autoencoders are neural network architectures designed to reconstruct their input by encoding it into a lower-dimensional representation and then decoding it \cite{pinaya2020autoencoders}. The reconstruction error, calculated as the Mean Squared Error (MSE) between the original input and its reconstruction, measures how well the model has learned normal patterns.

\begin{equation}\label{eq:mse}
    \text{MSE} = \frac{1}{n} \sum_{i=1}^{n} \left( x_i - \hat{x}_i \right)^2
\end{equation}

Higher reconstruction errors indicate sequences that deviate from expected behaviour, signalling potential anomalies. To classify sequences as anomalous, thresholds for reconstruction error were determined using a percentile-based approach, ranging from the 75th to 90th percentile. This approach balances anomaly detection precision and recall, capturing both subtle deviations and extreme outliers without overfitting to normal fluctuations. Sequences exceeding the threshold were flagged as anomalous, and contracts with more than 30\% anomalous transactions were classified as illicit..

By incorporating opcode embeddings alongside transactional features, the CAE captures both runtime anomalies and underlying structural issues. This integration improves detection robustness, distinguishing between benign transaction spikes and illicit activities. The multimodal approach enables a holistic analysis of smart contract reputability, addressing limitations of code-only or transactional-only methods and providing richer insights into potential vulnerabilities and behavioural anomalies.

\subsection{Evaluation Metrics}
We evaluate model performance using accuracy, recall, and F1-score, which provide complementary insights, particularly for imbalanced datasets. Recall highlights the model’s ability to detect illicit contracts, which is critical in this context to minimize missed detections of potentially fraudulent or malicious smart contracts. F1-score balances precision and recall, reflecting overall detection performance. These metrics are calculated using standard formulas and are referenced during results discussions.

\section{Experimentation and Results}
\subsection{AI-based Code Analysis for Reputability Prediction}

This section outlines the evaluation of boosting algorithms and data augmentation techniques for predicting smart contract reputability using code features. 

The initial LightGBM model trained on the original, imbalanced dataset failed to detect illicit smart contracts due to the disproportionate number of reputable smart contracts. To address this issue, augmentation techniques such as SMOTE and ADASYN were applied to the training set, increasing the number of illicit contracts from 132 to approximately 2,160 to balance the dataset. While this improved recall modestly, it also introduced more false positives, highlighting the difficulty of maintaining precision when working with imbalanced datasets.

GAN-based augmentation demonstrated the most notable improvement, surpassing SMOTE and ADASYN by generating realistic opcode embeddings that captured higher-order patterns. Table~\ref{tab:lgbm_performance_comparison} presents the detailed performance metrics for each dataset.

\begin{table}[htbp]
\caption{Performance of LightGBM on the Original and Augmented Datasets}
\begin{center}
\begin{tabular}{|c|c|c|c|}
\hline
\textbf{Dataset} & \multicolumn{1}{c|}{\textbf{Recall}} & \textbf{F1-Score} & \textbf{{Accuracy}} \\
& \multicolumn{1}{c|}{\textbf{(Illicit Class)}} & & \\
\hline
Original (No Augmentation) & 0.000 & 0.000 & 94.1\% \\
SMOTE & 0.431 & 0.427 & 93.1\% \\
ADASYN & 0.379 & 0.355 & 91.9\% \\
GAN-Augmented & 0.942 & 0.933 & 97.67\% \\
\hline
\multicolumn{4}{l}{$^{\mathrm{a}}$Evaluation metrics for predicting illicit smart contracts.}
\end{tabular}
\label{tab:lgbm_performance_comparison}
\end{center}
\end{table}

\begin{figure}[ht!]
    \centering
    \includegraphics[width=\linewidth]{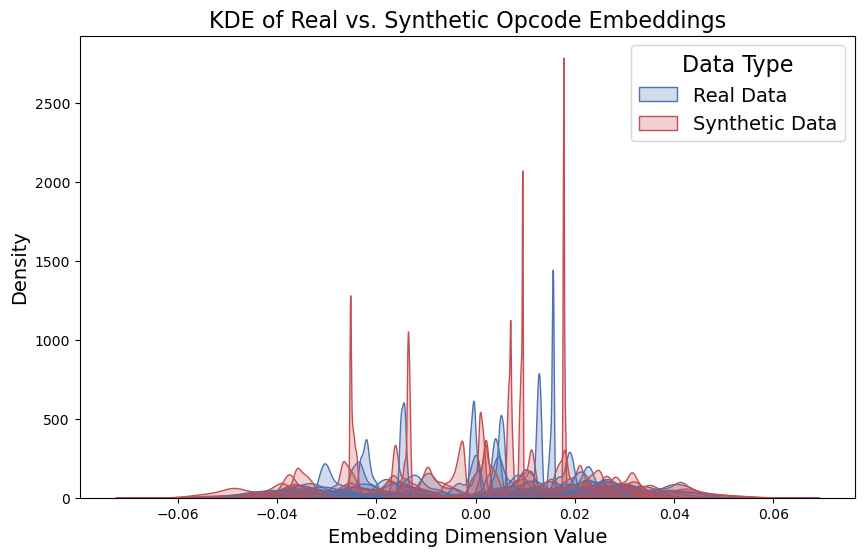}
    \caption{KDE plot comparing real and GAN-generated opcode embeddings.}
    \label{fig:kde_comparison}
\end{figure}

We also present the kernel density estimate (KDE) plot in Figure~\ref{fig:kde_comparison} which offers insight into the similarity between the two distributions. The substantial overlap between real and GAN-generated opcode embeddings, further validates the quality of synthetic data. Moreover, statistical metrics further support this, with a correlation coefficient of 0.946 and a variance ratio of 0.465, indicating that the GAN-generated embeddings closely align with the distribution of the original dataset while preserving realistic variability.

These findings highlight the limitations of SMOTE and ADASYN, which interpolate between existing samples, potentially missing complex patterns \cite{tanaka2019data}. In contrast, GAN-based augmentation captures nuanced relationships in the data, significantly improving the detection of illicit contracts.

Table~\ref{tab:boosting_performance} presents the performance metrics for LightGBM, CatBoost, and XGBoost models trained on augmented datasets. Notably, LightGBM achieved the highest recall (0.942) for detecting illicit smart contracts, underscoring its superior capacity to address class imbalance and capture complex feature interactions. High recall is particularly important in this context, as failing to detect illicit smart contracts poses significant security risks, potentially enabling fraudulent activities and undermining trust in decentralized ecosystems.

Among the models, LightGBM demonstrated the lowest log loss (0.051) across 5-fold cross-validation, indicating robust generalization and efficient learning from the augmented dataset. CatBoost, known for its automatic handling of categorical features and balanced accuracy, exhibited competitive performance with a slightly higher log loss (0.075). XGBoost, renowned for its adaptability and handling of sparse data, achieved recall parity with LightGBM but displayed a marginally higher log loss (0.053), likely due to its deeper tree structure. These results underscore LightGBM's superior balance between complexity and optimization.

\begin{table}[htbp]
\caption{Performance Metrics for Boosting Algorithms}
\begin{center}
\begin{tabular}{|l|c|c|c|c|}
\hline
\textbf{Algorithm} & \textbf{Accuracy} & \textbf{Recall} & \textbf{F1-Score} & \textbf{Log Loss} \\
 & & \textbf{(Illicit Class)} & & \\
\hline
LightGBM & 97.67\% & 0.942 & 0.933 & 0.051 \\
CatBoost & 97.58\% & 0.937 & 0.930 & 0.075 \\
XGBoost & 97.49\% & 0.942 & 0.928 & 0.053 \\
\hline
\end{tabular}
\label{tab:boosting_performance}
\end{center}
\end{table}

The hyperparameters used for each boosting algorithm, summarized in Table~\ref{tab:boosting_hyperparameters}, further illustrate why LightGBM excels in this context. LightGBM's smaller maximum depth (5) and higher number of estimators (300) allow it to build more shallow yet numerous trees, leading to smoother decision boundaries and reduced overfitting.

\begin{table}[htbp]
\caption{Optimal Hyperparameters for Boosting Algorithms}
\begin{center}
\begin{tabular}{|l|c|c|c|}
\hline
\textbf{Hyperparameter} & \textbf{LightGBM} & \textbf{XGBoost} & \textbf{CatBoost} \\
\hline
Learning Rate          & 0.1                & 0.2              & 0.1              \\
Maximum Depth          & 5                  & 8                & 8                 \\
Subsample              & 0.5                & 0.5              & 0.5                \\
Regularization Alpha   & 0.1                & 0.1              & 0.1                \\
Regularization Lambda  & 0.01               & 0.01             & 0.01                 \\
Number of Estimators   & 300                & 200              & 200               \\
\hline
\end{tabular}
\label{tab:boosting_hyperparameters}
\end{center}
\end{table}

To assess the generalizability of our model, we benchmarked it against the Ponzi dataset from \cite{Chen2018}, which comprises 3,769 contracts, including 200 labelled Ponzi schemes. Table~\ref{tab:benchmark_results} summarizes the comparison between our LightGBM model and the original XGBoost-based model from \cite{Chen2018}.

\begin{table}[htbp]
    \caption{Performance Comparison on the Ponzi Scheme Benchmark Dataset}
    \centering
    \begin{tabular}{|l|c|c|c|}
        \hline
        \textbf{Model} & \textbf{Recall} & \textbf{Precision} & \textbf{F1-Score} \\
        \hline
        LightGBM & 0.930 & 0.949 & 0.939 \\
        XGBoost \cite{Chen2018} & 0.810 & 0.940 & 0.860 \\
        \hline
    \end{tabular}
    \label{tab:benchmark_results}
\end{table}

The benchmark results in Table~\ref{tab:benchmark_results} demonstrate that our LightGBM model outperforms the XGBoost-based approach from \cite{Chen2018} in detecting Ponzi schemes. Specifically, LightGBM achieved higher recall and F1-score, indicating its improved ability to detect Ponzi schemes while maintaining a balanced trade-off between false positives and false negatives. The higher precision also suggests fewer false alarms compared to the benchmark model, reinforcing the robustness of GAN-augmented opcode embeddings in identifying illicit behaviour patterns with greater reliability.

Therefore, the results confirm that code features, such as opcode embeddings, provide significant insights into smart contract reputability, enabling accurate detection of illicit contracts. However, despite the improvements, code features alone may not fully capture behavioural shifts indicative of evolving reputability. This motivates the need for multimodal data fusion that integrates transactional data to enhance predictions, even in the absence of explicit reputability labels at specific timestamps.

\subsection{Multimodal Anomaly Detection}

The CAE model was first evaluated using transactional data in isolation. The dataset consisted of over 820,920 transactions from reputable smart contracts and 6,761 transactions from illicit contracts. Reconstruction errors served as the primary anomaly detection signal, where higher errors indicated greater deviation from normal behaviour. Since the CAE was trained exclusively on reputable contracts, it learned their underlying transactional patterns, resulting in lower reconstruction errors for reputable contracts. In contrast, illicit contracts, characterized by anomalous transactional behaviours, exhibited significantly higher reconstruction errors, reinforcing their deviation from normal patterns. This behaviour aligns with the hypothesis that higher reconstruction errors correspond to potentially non-reputable behaviour.

Despite this capability, as shown in Table~\ref{tab:performance_metrics_thresholds_combined}, even at the optimal threshold of 90, the transaction-only CAE achieved a recall of 90.79\% for the illicit class, with an F1-score of 0.920. However, it was prone to false positives during legitimate transaction surges, highlighting the limitations of relying solely on transactional data. The threshold-setting process demonstrated that while higher thresholds reduced false positives, they also risked omitting true anomalies, creating a trade-off between sensitivity and specificity.

\begin{figure}[ht!]
\centering
\includegraphics[width=\linewidth]{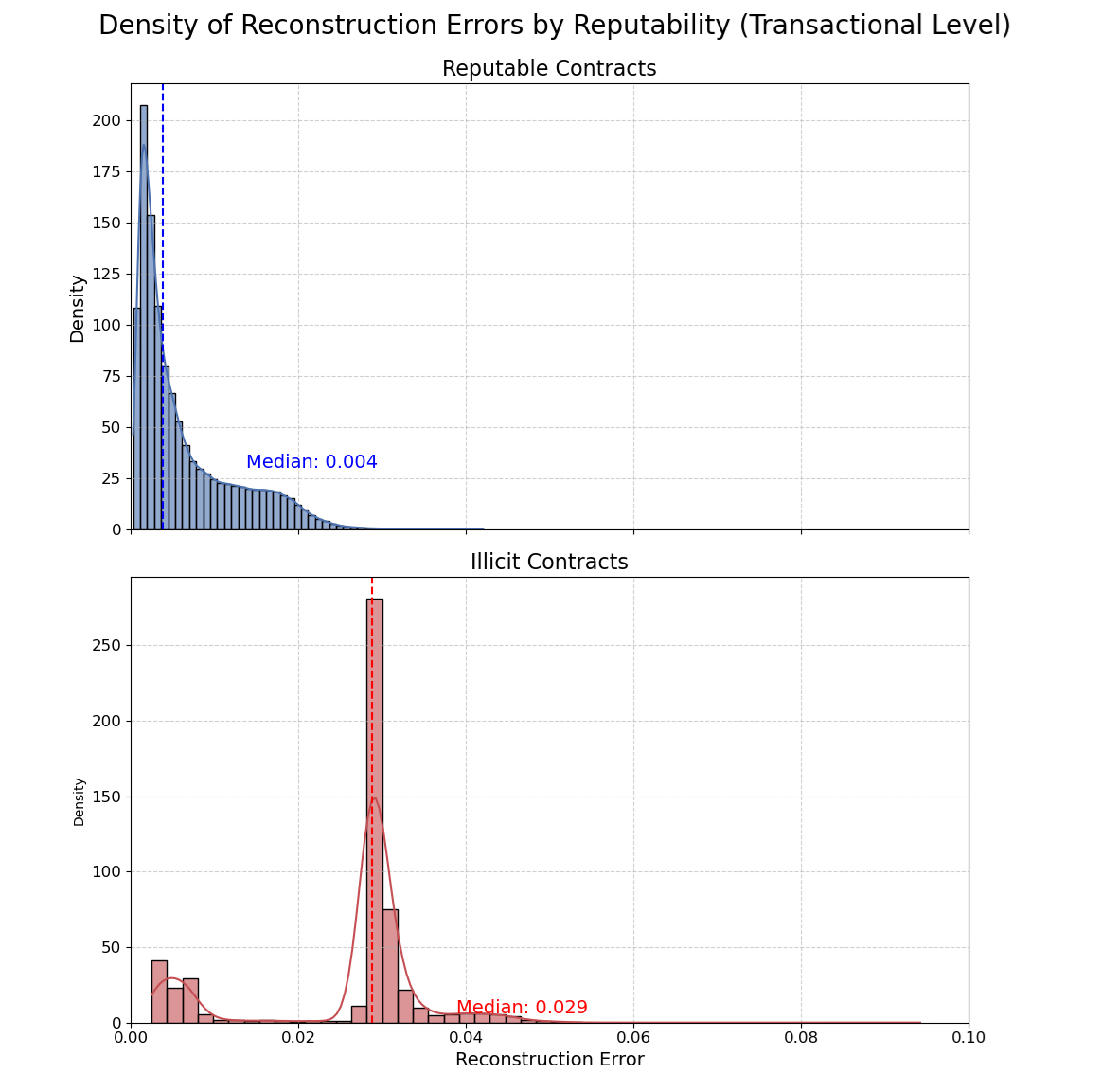}
\caption{Reconstruction Error Distribution for the Multimodal CAE.}
\label{fig:multimodal_reconstruction_error_distribution}
\end{figure}

To address these limitations, we developed a multimodal CAE that integrates opcode embeddings with transactional features. This multimodal approach incorporates code and dynamic data, enabling the model to capture both structural vulnerabilities and behavioural anomalies. As illustrated in Figure~\ref{fig:multimodal_reconstruction_error_distribution}, the reconstruction error distribution for reputable contracts narrowed, with the median error for reputable contracts decreasing significantly compared to the transaction-only model. This indicates that the inclusion of opcode embeddings enhanced the model’s ability to capture the nuances of normal smart contract behaviour. In contrast, reconstruction errors for illicit contracts remained notably high, emphasizing their anomalous nature and reinforcing the importance of opcode embeddings in detecting complex irregularities within smart contract behaviours.

\begin{table*}[ht!]
    \caption{Performance Metrics for Transaction-Only and Multimodal CAEs Across Different Thresholds}
    \centering
    \begin{tabular}{|c|c|c|c|c|c|}
        \hline
        \textbf{Model} & \textbf{Threshold} & \textbf{Accuracy} & \textbf{Recall (Illicit)} & \textbf{F1-Score (Reputable)} & \textbf{F1-Score (Illicit)} \\
        \hline
        \multirow{4}{*}{Transaction-Only CAE} 
        & 75  & 0.916  & 0.882  & 0.946  & 0.812  \\
        & 80  & 0.940  & 0.855  & 0.962  & 0.855  \\
        & 85  & 0.956  & 0.855  & 0.973  & 0.890  \\
        & \textbf{90}  & \textbf{0.967}  & \textbf{0.908}  & \textbf{0.980}  & \textbf{0.920}  \\
        \hline
        \multirow{4}{*}{Multimodal CAE} 
        & 75  & 0.926  & 0.895  & 0.953  & 0.834  \\
        & 80  & 0.943  & 0.882  & 0.964  & 0.865  \\
        & 85  & 0.965  & 0.882  & 0.978  & 0.912  \\
        & \textbf{90}  & \textbf{0.992}  & \textbf{0.974}  & \textbf{0.995}  & \textbf{0.980}  \\
        \hline
    \end{tabular}
    \label{tab:performance_metrics_thresholds_combined}
\end{table*}

The performance comparison across different thresholds for the transaction-only and multimodal CAEs is summarized in Table~\ref{tab:performance_metrics_thresholds_combined}. At the optimal threshold of 90, the multimodal CAE improved recall for illicit contracts by 7.25\%, demonstrating the benefits of integrating transactional data with code features. This threshold was identified as optimal because it achieved the best trade-off between precision and recall, minimizing False Positives (FP) which are illicit contracts incorrectly classified as reputable—while maintaining high detection rates for illicit contracts.

This enhancement reduced FP and allowed the model to better distinguish true anomalies from benign deviations. As shown in Table~\ref{tab:performance_metrics_thresholds_combined}, the inclusion of opcode embeddings provided richer contextual information, highlighting the effectiveness of multimodal fusion for robust anomaly detection, particularly in scenarios where code features alone are insufficient for capturing reputability shifts.

\begin{figure}[ht!] \centering \includegraphics[width=1.0\linewidth]{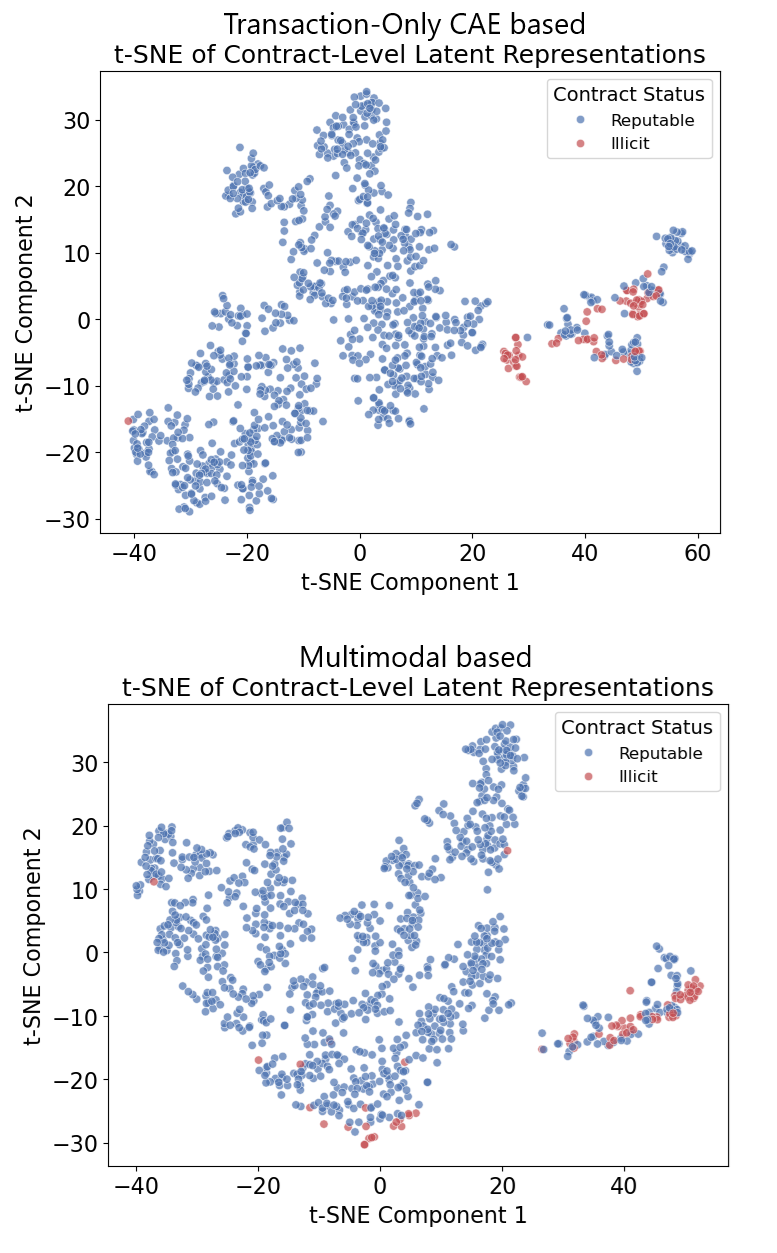} \caption{t-SNE visualizations of contract-level latent representations for the Transaction-Only CAE (top) and Multimodal CAE (bottom). The Transaction-Only CAE shows some overlap between reputable (blue) and illicit (red) contracts, while the Multimodal CAE forms more distinct clusters, indicating improved separation with multimodal data fusion.} \label{fig:tsne_pca_compare} \end{figure}

To further validate the performance improvements of the multimodal CAE, latent space visualizations were generated using t-SNE \cite{van2008visualizing} as shown in Figure~\ref{fig:tsne_pca_compare}. These plots illustrate the clustering of reputable and illicit contracts in both the transaction-only and multimodal models. In the transaction-only CAE, there was noticeable overlap between reputable and illicit contracts, particularly during benign but atypical transaction surges. In contrast, the multimodal CAE formed more distinct clusters, with minimal overlap, demonstrating its enhanced ability to separate reputable and illicit contracts.

This enhanced separation in the latent space underscores the contribution of opcode embeddings in reinforcing class separability. By leveraging both code and dynamic features, the multimodal CAE learns richer latent representations that effectively capture behavioural and structural nuances in smart contract transactions.

In summary, the multimodal CAE outperforms the transaction-only CAE across all key metrics, achieving a recall of 0.974 and an F1-score of 0.980 for illicit contracts at the optimal threshold of 90. These results demonstrate the significant advantage of integrating opcode embeddings with transactional data for anomaly detection. The findings support the hypothesis that multimodal data fusion enhances smart contract reputability prediction by capturing both structural vulnerabilities and behavioural anomalies, demonstrating the value of combining AI-based code analysis with dynamic transactional data for a comprehensive reputability assessment.

\section{Limitations and Future Directions}

This study provided valuable insights into smart contract reputability prediction and anomaly detection, demonstrating the advantages of multimodal data fusion. However, several limitations were encountered, which highlight opportunities for future exploration.

A key challenge in this work was the data imbalance between reputable and illicit smart contracts. Reputable contracts were more prevalent and contained a significantly higher average number of transactions, resulting in a skewed dataset. While data augmentation techniques such as SMOTE, ADASYN, and GAN-based augmentation were applied, synthetic data may not fully capture the diversity of real-world illicit contract behaviours. This raises the need for anomaly detection techniques that can perform robustly despite limited samples of illicit contracts. Future efforts could focus on acquiring richer datasets or developing more advanced augmentation methods capable of introducing variability without overfitting to known patterns.

Another limitation stems from the lack of explicit reputability labels at specific timestamps, necessitating the use of proxy indicators such as reconstruction errors and reputability scores derived from model predictions. Although this approach provided valuable insights, it introduces ambiguity, as proxy scores may not perfectly align with real-world reputability dynamics, particularly during subtle or irregular behavioural shifts. This ambiguity can affect interpretability and model robustness, particularly in edge cases where reputability changes are gradual or inconsistent. Further research could integrate additional contextual signals, such as smart contract event logs, user interactions, or expert annotations, to enhance reputability tracking accuracy and reduce reliance on indirect indicators.

The complexity of multimodal fusion also posed challenges. Integrating code and transactional features increased the computational complexity of the model and required extensive hyperparameter tuning to balance the representation of both data modalities. Computational constraints limited the scope of tuning experiments, potentially impacting full optimization. Moreover, the resource demands of the model may hinder real-time analysis across large blockchain networks. Addressing these challenges may involve developing more efficient fusion frameworks, such as lightweight architectures or distributed processing methods, to enhance scalability while preserving the benefits of multimodal integration.

Additionally, this study did not fully address the temporal evolution of reputability across longer time windows. While the multimodal CAE demonstrated strong performance in detecting reputability shifts within short-term windows, smart contracts may experience gradual changes that require advanced temporal modelling frameworks to detect long-term reputational risks. Incorporating sequential data fusion and time-series models could enhance the framework’s ability to capture evolving behavioural patterns over extended periods.

By addressing these limitations, future research can further improve the robustness, interpretability, and scalability of anomaly detection in decentralized systems. This work underscores the importance of multimodal approaches in capturing complex blockchain behaviours and sets the foundation for more advanced frameworks that adapt to the dynamic nature of smart contract ecosystems.

\section{Concluding Remarks}
This study demonstrated the effectiveness of leveraging AI-based code analysis for smart contract reputability prediction, showing that opcode embeddings provide valuable insights into contract vulnerabilities and reputability trends. By employing boosting algorithms and anomaly detection techniques, the framework accurately identified illicit contracts, reinforcing the role of code-based features in distinguishing reputable behaviour. However, code features alone struggle to capture evolving behavioural shifts, highlighting the need for a more comprehensive approach that considers dynamic interactions and real-time patterns.

To address this, a multimodal data fusion framework was developed, integrating transactional data with code features to capture both structural and behavioural anomalies. The inclusion of transactional data significantly enhanced the model’s ability to distinguish true anomalies from benign deviations, demonstrating the value of combining code and dynamic information for a more holistic reputability assessment. This capability is particularly relevant for DeFi platforms, where smart contracts facilitate critical processes such as lending, staking, and token exchanges, often handling substantial financial transactions autonomously. In such high-stakes environments, the early detection of anomalous contract behaviour is crucial for preserving user trust and minimizing potential financial losses. The ability to detect illicit activity in near real-time not only strengthens the resilience of DeFi ecosystems but also supports regulatory compliance efforts by providing actionable insights into contract performance and reputability trends.

% Future extensions of this research could focus on enhancing the system's responsiveness by incorporating real-time data streams and more advanced attention-based fusion mechanisms. This would enable the framework to dynamically prioritize features based on contextual relevance, further improving its ability to capture nuanced patterns in smart contract behaviour. Additionally, expanding the dataset to include a wider variety of smart contracts, particularly those involved in more complex interactions, could improve the model’s robustness and generalizability. By continuing to refine and expand multimodal approaches, this study provides a foundational step toward more secure, transparent, and trustworthy blockchain ecosystems.

Future extensions of this research could focus on enhancing the system's responsiveness by incorporating real-time data streams and more advanced attention-based fusion mechanisms. This would enable dynamic feature prioritization based on contextual relevance, further improving the capture of nuanced smart contract behaviour. Additionally, expanding the dataset to include a broader range of smart contracts, particularly those involved in complex interactions, could strengthen the model’s robustness and generalizability. By refining and expanding multimodal approaches, this study provides a foundational step toward more secure, transparent, and trustworthy blockchain ecosystems.

\bibliographystyle{IEEEtran}
\bibliography{references}
% \begin{thebibliography}{00}
% \bibitem{b1} G. Eason, B. Noble, and I. N. Sneddon, ``On certain integrals of Lipschitz-Hankel type involving products of Bessel functions,'' Phil. Trans. Roy. Soc. London, vol. A247, pp. 529--551, April 1955.
% \bibitem{b2} J. Clerk Maxwell, A Treatise on Electricity and Magnetism, 3rd ed., vol. 2. Oxford: Clarendon, 1892, pp.68--73.
% \bibitem{b3} I. S. Jacobs and C. P. Bean, ``Fine particles, thin films and exchange anisotropy,'' in Magnetism, vol. III, G. T. Rado and H. Suhl, Eds. New York: Academic, 1963, pp. 271--350.
% \bibitem{b4} K. Elissa, ``Title of paper if known,'' unpublished.
% \bibitem{b5} R. Nicole, ``Title of paper with only first word capitalized,'' J. Name Stand. Abbrev., in press.
% \bibitem{b6} Y. Yorozu, M. Hirano, K. Oka, and Y. Tagawa, ``Electron spectroscopy studies on magneto-optical media and plastic substrate interface,'' IEEE Transl. J. Magn. Japan, vol. 2, pp. 740--741, August 1987 [Digests 9th Annual Conf. Magnetics Japan, p. 301, 1982].
% \bibitem{b7} M. Young, The Technical Writer's Handbook. Mill Valley, CA: University Science, 1989.
% \end{thebibliography}
% \vspace{12pt}
% \color{red}
% IEEE conference templates contain guidance text for composing and formatting conference papers. Please ensure that all template text is removed from your conference paper prior to submission to the conference. Failure to remove the template text from your paper may result in your paper not being published.

\end{document}